\documentclass[lettersize,journal]{IEEEtran}
\usepackage{cite}
\usepackage{multirow}
\usepackage{amsmath}
\usepackage{color}
\usepackage{xcolor}
\usepackage{amssymb}
\usepackage{multirow}
\usepackage{float}
\usepackage{verbatim}
\usepackage{diagbox}
\usepackage{rotating}
\usepackage{booktabs}
\usepackage[detect-all]{siunitx}
\usepackage[hidelinks]{hyperref}
\usepackage{caption}
\usepackage{subcaption}
\usepackage{tikz}
\usepackage{stfloats}
\usetikzlibrary{spy}
 \usepackage{algorithm}
 \usepackage{algorithmic}
\hypersetup{
  colorlinks=true,%
  urlcolor=magenta
}

\definecolor{dgreen}{rgb}{0, 0.6, 0} 
\definecolor{cyan}{rgb}{0, 0.5, 0.6} 
\definecolor{Darkviolet}{rgb}{0.58, 0, 0.83}

\newcommand{\etal}{{\it et al.}}
  
\newcommand{\heading}[1]{\noindent\textbf{#1}}

\begin{document}

\title{Certified Zeroth-order Black-Box Defense with Robust UNet Denoiser}
 \author{Astha Verma
IIIT Delhi
 {\tt\small asthav@iiitd.ac.in}\\
 \and
A V Subramanyam
 IIIT Delhi 
 {\tt\small subramanyam@iiitd.ac.in}\\
  \and
Siddhesh Bangar
MIDAS LAB, IIIT Delhi 
 {\tt\small siddheshb008@gmail.com}\\
 \and
Naman Lal
 MIDAS LAB, IIIT Delhi 
 {\tt\small namanlal.lal92@gmail.com}\\
    \and
Rajiv Ratn Shah
 IIIT Delhi 
 {\tt\small rajivratn@iiitd.ac.in}\\
 \and
 Shin'ichi Satoh
 NII Tokyo 
 {\tt\small satoh@nii.ac.jp}\\
 }

\maketitle

\begin{abstract}
   Certified defense methods against adversarial perturbations have been recently investigated in the black-box setting with a zeroth-order (ZO) perspective. However, these methods suffer from high model variance with low performance on high-dimensional datasets due to the ineffective design of the denoiser and are limited in their utilization of ZO techniques. To this end, we propose a certified ZO preprocessing technique for removing adversarial perturbations from the attacked image in the black-box setting using only model queries. We propose a robust UNet denoiser (RDUNet) that ensures the robustness of black-box models trained on high-dimensional datasets. We propose a novel black-box denoised smoothing (DS) defense mechanism, ZO-RUDS, by prepending our RDUNet to the black-box model, ensuring black-box defense. We further propose ZO-AE-RUDS in which RDUNet followed by autoencoder (AE) is prepended to the black-box model. We perform extensive experiments on four classification datasets, CIFAR-10, CIFAR-100, Tiny Imagenet, STL-10, and the MNIST dataset for image reconstruction tasks. Our proposed defense methods ZO-RUDS and ZO-AE-RUDS beat SOTA with a huge margin of $35\%$ and $9\%$, for low dimensional (CIFAR-10) and with a margin of $20.61\%$ and $23.51\%$ for high-dimensional (STL-10) datasets, respectively.
\end{abstract}
\section{Introduction}
A notable amount of success has been attained by machine learning (ML) models~\cite{joshi2022certified, katz2022formal}, and deep neural networks (DNNs) in particular because of their better predictive capabilities. However, their lack of robustness and susceptibility to adversarial perturbations has caused serious worries about their wide-scale adaptation in a number of artificial intelligence (AI) applications~\cite{goodfellow2014explaining, carlini2017towards, papernot2016limitations, brown2017adversarial, eykholt2018robust, antun2020instabilities}. These adversarial attacks have motivated various strategies to strengthen ML models as a key area of research~\cite{madry2017towards,athalye2018obfuscated,zhang2019theoretically,cui2021learnable,verma2022wasserstein}. Among these techniques, adversarial training (AT)~\cite{szegedy2013intriguing,madry2017towards} is one of the prominent defense strategies. The advancements in AT led to various empirical defense methods~\cite{athalye2018synthesizing,wang2022self,yan2022wavelet,chan2019jacobian,zhang2019theoretically,verma2023meta}, however these methods may not always be certifiably robust~\cite{pmlr-v80-uesato18a,10.5555/3524938.3525144}. Another line of research is certified defense, where an off-the-shelf model's prediction is certified within the neighborhood of the input. These methods are called certified defense techniques~\cite{wong2018provable, raghunathan2018certified, katz2017reluplex, salman2019provably, salman2020denoised, salman2022certified}.
\begin{figure}[t]
\centering
\includegraphics[width=0.9\linewidth]{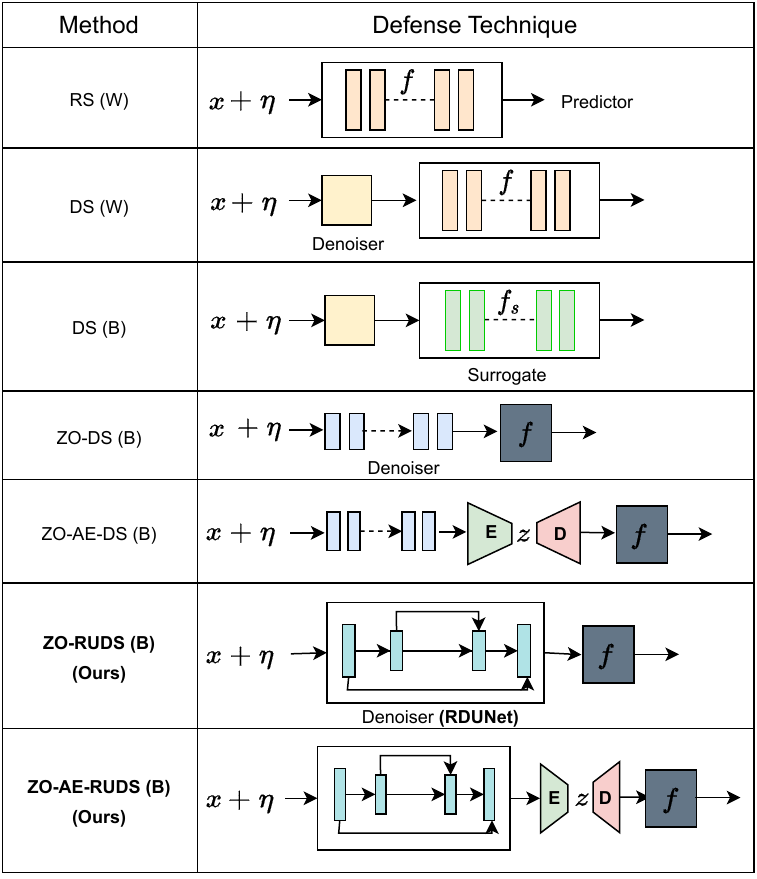}
\caption{We make a comparison with four previous certified defense methods, including RS~\cite{cohen2019certified}, DS (W)~\cite{salman2020denoised},  DS (B)~\cite{salman2020denoised}, ZO-DS~\cite{zhang2022robustify} and ZO-AE-DS~\cite{zhang2022robustify} (ZO-optimization approaches). `W' and `B' refer to white-box (defense technique can utilize weights of target model $f$) and black-box settings. `$x$' - input sample, `$\eta$' - noise, `E' - Encoder, `D' - Decoder, `$f$' - target model, $f_{s}$ - surrogate model (proxy of $f$), and `$z$' - latent feature vector. }
\label{fig:fig1}
\end{figure}

Cohen~\etal~\cite{cohen2019certified} first proposed randomized smoothing (RS), which certifies defense by forming a smoothed model from the empirical model by adding Gaussian noise to the input images. Few other works have been proposed which provide certified defense inspired by randomized smoothing~\cite{salman2020denoised,salman2022certified,addepalli2021boosting}. Salman~\etal~\cite{salman2020denoised} pre-pended a custom-trained denoiser to the predictor for increased robustness. In another work, Salman~\etal~\cite{salman2022certified} apply visual transformers within the smoothing network in order to provide certified robustness to adversarial patches. While previous works in adversarial defense have achieved promising advancements, robustness is provided over white-box models with known architectures and parameters. The white-box assumption, however, has high computational complexity as models are trained end-to-end as in AT, thus limiting the practicality and scalability of the defense method. For instance, it becomes impractical to retrain complex ML models trained on a vast number of MRIs or CT scans~\cite{sinha2022ml}~\cite{hussain2023deep}. 

Moreover, privacy concerns may arise when implementing white-box defense since the owner may not wish to reveal model information. This is because attacks such as membership inference and model inversion attacks expose the vulnerabilities of the training data~\cite{fredrikson2015model}. Due to the scalability and privacy issues, few previous works tackled the highly non-trivial problem of adversarial defense in the black-box setting (`black-box defense')~\cite{salman2020denoised,zhang2022robustify}.

Salman~\etal~\cite{salman2020denoised} used surrogate models as approximates of the black-box models, over which defense may be done using the white-box setup. However, this setup requires information on the target model type and its function, which may be not be available practically. In another recent work, Zhang~\etal~\cite{zhang2022robustify} proposed a more authentic black-box defense of DNN models with the help of the zeroth-order optimization perspective. They pre-pended a custom-trained denoiser as in~\cite{salman2020denoised} followed by an autoencoder architecture to the target model and trained it with ZO optimization. However,~\cite{zhang2022robustify} fails to perform for high dimensional datasets like Tiny imagenet with the dimensionality of $299 \times 299 \times 3$ as the denoiser used may fail to preserve spatial information. The main contribution of ~\cite{zhang2022robustify}, that is, the addition of autoencoder to existing technique~\cite{salman2020denoised}, enhances the robustness of black-box model to some extent for low-dimensional datasets like CIFAR-10 ($32 \times 32 \times 3$). However, this limits its usage to only the coordinate-wise gradient estimation (CGE) ZO optimization technique. However, our proposed approach can utilize two main existing ZO optimization techniques: randomized gradient estimate (RGE) and CGE. We discuss in detail in subsection~\ref{subsec:Image}, where we prove the limitations of~\cite{zhang2022robustify} on high dimensional datasets graphically. 

 We propose a certified black-box defense ZO-RUDS using the ZO optimization technique, where we pre-pend a novel robust UNet denoiser (RDUNet) to the target model. We further pre-pend our RDUNet and custom-trained autoencoder and propose the ZO-AE-RUDS defense mechanism. Our proposed methods require only input queries and output feedback and provide defense in a pure black-box setting. Unlike SOTA~\cite{zhang2022robustify}, which leads to high model variance on direct application of ZO optimization on the custom-trained denoiser, our proposed RDUNet, due to its architectural advantage over previous denoisers decreases model variance and provides better performance with direct application of ZO optimization. We experiment with various denoisers and prove that our RDUNet denoiser provides improved performance for both low-dimensional and high-dimensional datasets. Since we are dealing with a difficult ZO optimization and cannot back-propagate through the model, we optimize our proposed model by utilizing the black-box model's predicted labels and softmax probabilities. In order to further increase the certification of our proposed approach, we utilize maximum mean discrepancy (MMD) to bring the distributions of original input images closer to obtained denoised output. We provide an illustration of certified defense techniques and compare them to our approach in Figure~\ref{fig:fig1}. 

In Figure~\ref{fig:fig1}, we compare our proposed approaches, ZO-RUDS and ZO-AE-RUDS, with SOTA certified defense techniques in white-box (W) and black-box (B) settings. The input to the defense framework is sample $x$ and noise $\eta$. RS~\cite{cohen2019certified} and DS~\cite{salman2020denoised} provide certified defense in the white-box setting. In the `RS' technique, noisy images ($x+\eta$) are input to the white-box model for certified robustness. `DS' uses an additional custom-trained denoiser and pre-pends it to the predictor for certified robustness in the white-box setting. In addition to defense in the white-box setting, `DS' proposed certified black-box robustness using a surrogate model.

In order to provide black-box defense without the use of a surrogate model ( as it uses the target model as its proxy and it is not always possible to have access to the information on the target model and its function), Zhang~\etal~\cite{zhang2022robustify} proposed black-box defense with zeroth-order (ZO) optimization. They proposed ZO-DS by direct application of ZO on `DS' and further append an autoencoder in the ZO-AE-DS technique. However, their proposed techniques have low performance on high-dimension datasets as the custom-trained denoiser fails to learn fine-scale information leading to poor performance. In order to overcome these limitations, we propose a robust UNet denoiser RDUNet inspired from the conventional UNet used for image segmentation~\cite{ronneberger2015u}. Our robust UNet RDUNet with the upsampling and downsampling layers, and lateral skip connections enables our defense to learn complex structures and fine-scale information and makes it invariant to changes in image dimensions, thus giving a high performance for high-dimension images.
 
We summarize our contributions as follows:
\begin{itemize}
    \item We propose a certified black-box defense mechanism based on the preprocessing technique of pre-pending a robust denoiser to the predictor to remove adversarial noise using only the input queries and the feedback obtained from the model.
    \item We design a novel robust UNet denoiser RDUNet which defends a black-box model with ZO optimization approaches. Unlike previous ZO optimization-based defense approaches, which give a poor performance on high-dimensional data due to high model variance, our UNet-based robustification model gives high performance for both low-dimensional and high-dimensional datasets. 
    \item We conduct extensive experiments and show that our proposed defense mechanism beats SOTA by a huge margin on four classification datasets, CIFAR-10, CIFAR-100, STL-10, Tiny Imagenet, and on MNIST dataset for reconstruction task.
\end{itemize}

\section{Related Work}
We broadly categorize the literature on robust defense into empirical and certified defense and briefly discuss them below in the white-box and black-box settings.
\subsection{Empirical Defense}
Szegdy~\etal~\cite{szegedy2013intriguing} first proposed robust empirical defense in the form of adversarial training (AT). Due to AT, there has been a rapid increase in empirical defense methods~\cite{wang2022self, yan2022wavelet,cheng2023adversarial,wei2023self,chan2019jacobian}. Zhang~\etal~\cite{zhang2019theoretically} proposed a tradeoff between robustness and accuracy that can optimize defense performance. In order to improve the scalability of AT, empirical robustness is provided by previous works which design computationally light alternatives of AT~\cite{carmon2019unlabeled,sehwag2021robust}. Some recent empirical defense works are based on the concept of distillation, initially proposed by Hinton~\etal~\cite{hinton2015distilling}. Papernot~\etal~\cite{papernot2016distillation} presented a defensive distillation strategy to counter adversarial attacks. Folz~\etal~\cite{folz2020adversarial} gave a distillation model for the original model, which is trained using a distillation algorithm. It masks the model gradient in order to prevent adversarial perturbations from attacking the model's gradient information. Addepalli~\etal~\cite{addepalli2020towards} proposed unique bit plane feature consistency (BPFC) regularizer to increase the model's resistance to adversarial attacks.
\subsection{Certified Defense}
Unlike empirical defense, the certified defense provides formal verification of robustness of the DNN model~\cite{ma2021towards, gupta2021verifiable, elaalami2022bod}. Certified robustness is given by a `safe' neighbourhood region around the input sample where the prediction of DNN reamins same. Previous works~\cite{katz2017reluplex,tjeng2017evaluating,bunel2018unified,dutta2017output} in this field provide `exact' certification, which is often compute-intensive and is not scalable to large architectures. Katz~\etal~\cite{katz2017reluplex} proposed a robust simplex verification method to handle non-linear ReLU activation functions. Another line of work~\cite{wong2018provable,zhang2018efficient}, which provides `incomplete' verification, requires less computation; however, it gives faulty certification and can decline certification even in the absence of adversarial perturbation. Both `exact' and `incomplete' posthoc certification methods require customized architectures and hence are not suitable for DNNs~\cite{zhang2022robustify}.

Another area of study focuses on in-process certification-aware training and prediction. For instance, a randomized smoothing (RS) involves perturbing the input samples with Gaussian noise. This process allows for the transformation of a given base classifier $f$ into a new ``smoothed classifier" $g$, using randomized smoothing. Importantly, this transformation ensures that $g$ is certified to be robust in the $L_{2}$ norm. In~\cite{cohen2019certified}, it was demonstrated that RS could offer formal assurances for adversarial robustness. As well as, there are several different RS-oriented verifiable defences that have been developed, including adversarial smoothing~\cite{salman2019provably}, denoised smoothing~\cite{salman2020denoised}, smoothed ViT~\cite{salman2022certified}, and feature smoothing~\cite{addepalli2021boosting}. 

\subsection{ZO Optimization for adversarial learning.} 
ZO optimization is useful in solving black-box problems where gradients are difficult to compute or infeasible to obtain~\cite{yin2023generalizable, wei2022simultaneously}. These methods are gradient-free counterparts of first-order (FO) optimization methods~\cite{liu2020primer}. Recently, ZO optimization has been used for generating adversarial perturbations in black-box setting~\cite{chen2017zoo,ilyas2018prior, ilyas2018black, tu2019autozoom, liu2019signsgd, liu2020min, huang2020black, cai2021zeroth, cai2022zeroth}. Similar to attack methods, ZO optimization can also be applied to black-box defense methods with access only to the inputs and outputs of the targeted model. Zhang~\etal~\cite{zhang2022robustify} proposed black-box defense using ZO optimization and leveraged autoencoder architecture for optimizing the defense approach with CGE optimization. However, their approach fails to perform for high-dimension datasets. Inspired from~\cite{zhang2022robustify}, we propose a better defense mechanism with a robust UNet denoiser which gives high performance for high-dimension images.

\section{Preliminaries} Let $x \in \mathbb{R}^{d}$ is the input sample and $l \in \{1, 2,....,Y\}$ be the label. An adversarial attack can perturb $x$ by adding an adversarial noise. In order to defend model $f$ against these adversarial attacks and to provide certified robustness,~\cite{cohen2019certified}~proposed randomized smoothing (RS), a technique to construct a smoothed classifier $f_{s}$ from $f$. It is given as,
\begin{equation}
\begin{aligned}
f_{s}(x) = \underset{l\in Y}{\arg \max} \; \mathbb{P}_{\eta \in N (0, \sigma^2 I)} [f(x + \eta) = l],
\label{eqn:Equation1}
\end{aligned}
\end{equation}
where $\mathbb{P}$ is the probability function, and $\eta$ is the Gaussian noise with standard deviation $\sigma$.

\heading{Randomized Smoothing and Certified Robustness.}
~\cite{lecuyer2019certified} and~\cite{li2019certified} first gave robustness guarantees for the smoothed classifier $f_{s}$ using RS, but it was loosely bounded.~\cite{cohen2019certified} proposed a tight bound on $l_{2}$ robustness guarantee for the smoothed classifier $f_{s}$. They used Monte Carlo sampling and proposed an effective statistical formulation for predicting and certifying $f_{s}$. If the prediction of the base classifier for noise perturbed input samples $(x+\eta)$ is the probability $p_{f}$ as the topmost prediction, and $p_{s}$ is the runner-up prediction, then the smoothed classifier is robust within the radius $R_{c}$ assuming that $f_{s}$ gives correct prediction. $R_{c}$ is the certified radius within which the predictions are guaranteed to remain constant. ~\cite{cohen2019certified} gave lower and upper bound estimates for $p_{f}$ and $p_{s}$ as $\underline{p_{f}}$ and $\overline{p_{s}}$ respectively using Monte Carlo technique~\cite{cohen2019certified}.

Given f is the base classifier which returns the target label of the input sample, and $f_{s}$ is the smoothed classifier, then assuming that $f_{s}$ classifies correctly, the probabilities of topmost and runner-up predictions are given as:

\begin{equation}
p_{f} = {max} \; \mathbb{P}_{\eta}[f(x + \eta) = l]
\label{eqn:Equation2}
\end{equation}

\begin{equation}
    p_{s} = \underset{l' \neq l}{max} \; \mathbb{P}_{\eta}[f(x + \eta) = l'],
    \label{eqn:Equation3}
\end{equation}

where $\eta$ is the noise sampled from the Gaussian distribution $N(0,\sigma^{2})$. Then, $f_{s}$ is robust inside a radius $R_{c}$, which is given as:

\begin{equation}
R_{c} = \frac{\sigma}{2}[\phi^{-1} (p_{f}) - \phi^{-1} (p_{s})],
\label{eqn:Equation4}
\end{equation}
where $\phi^{-1}$ is the inverse of standard Gaussian CDF. If $p_{f}$ and $p_{s}$ hold the below inequality:
\begin{equation}
    p_{f} \geq \underline{p_{f}} \geq \overline{p_{s}} \geq p_{s}
    \label{eqn:Equation5}
\end{equation}
Then, $R_{c}$ is given as:
\begin{equation}
    R_{c} = \frac{\sigma}{2}[\phi^{-1} (\underline{p_{f}}) - \phi^{-1} (\overline{p_{s}})]
    \label{eqn:Equation6}
\end{equation}

The enhancement of the smoothed classifier's robustness relies on the application of the ``Neyman-Pearson" lemma, which is used to derive the above expressions~\cite{cohen2019certified}.

\heading{Denoised Smoothing (DS).}~\cite{salman2020denoised} proposed that naively applying randomized smoothing gives low robustness, as the standard classifiers are not trained to be robust to the Gaussian perturbation of the input sample. `DS' augments a custom-trained denoiser $D_{s}^{\Theta}$ to base classifier $f$. In this approach, an image-denoising pre-processing step is employed before input samples are passed through $f$. The denoiser pre-pended smoothed classifier, which is effective at removing the Gaussian noise, is given as,
\begin{equation}
    f_{s}(x) = \underset{l \in \mathbb{Y}}{\arg \max} \; \mathbb{P}_{\eta \in N (0, \sigma^2 I)} [f(D_{s}^{\Theta}(x + \eta)) = l]. 
    \label{eqn:Equation7}
\end{equation}
In order to obtain the optimal denoiser $D_{s}^{\Theta}$, DS proposed a stability regularized denoising loss in the first-order optimization setting. It is given as,
\begin{equation}
\begin{split}
    \mathcal{L}_{MSE+STAB}(\Theta) = & \underset{T, \eta}{\mathbb{E}} \|D_{s}^{\Theta}(x + \eta) - x \|^{2}_{2} \\ 
    & + \underset{T, \eta}{\mathbb{E}}[\mathcal{L}_{CE}(f(D_{s}^{\Theta}(x + \eta)), f(x))],
    \end{split}
    \label{eqn:Equation8}
\end{equation}
where $T$ is the training dataset and $\mathcal{L}_{CE}$ is the cross-entropy loss. However, previous approaches provide certified robustness in white-box setting with access to the target model's architectures and parameters.~\cite{salman2020denoised} first proposed certified defense in black-box setting. However, they utilized surrogate model which requires the information about model type and its function. Recently,~\cite{zhang2022robustify} proposed certified black-box defense with ZO optimization.

\section{Methodology}
\label{sec:Method}
\begin{figure*}[h]
\centering
\includegraphics[width=0.9\linewidth]{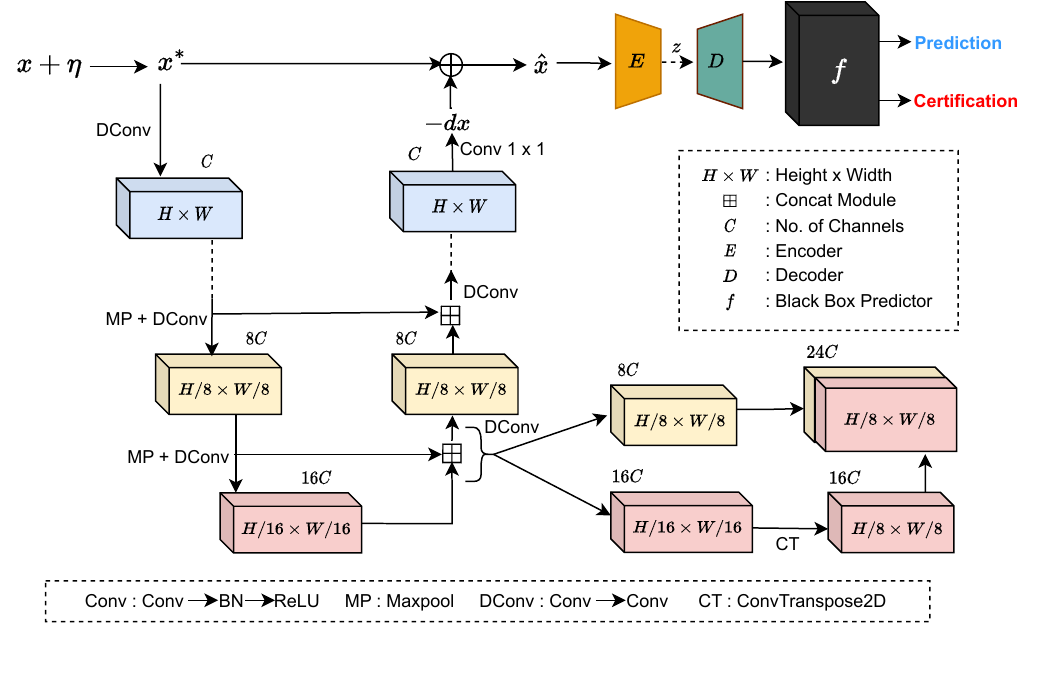}
\caption{An overview of proposed certified defense mechanism via robust UNet denoiser RDUNet. Noise $\eta$ is added to input sample $x$ which is given as input to the robust denoiser. The output of denoiser is the residual map which when added to noisy image $x^{\ast}$ gives denoised output $\hat{x}$. The denoised output is input of the autoencoder architecture which is then send as input to black-box model $f$.} 
\label{fig:fig_arch}
\end{figure*}

In this section, we first describe the architecture of our proposed robust defense model. We then describe the objective function of our proposed defense mechanism. Lastly, we discuss our two proposed defense mechanisms using RGE and CGE ZO optimization approaches. Our proposed framework is as shown in Figure~\ref{fig:fig_arch}.

\heading{Problem Statement.} We aim to defend a black-box model $f$, where $f$ is used for classification or reconstruction purposes. We consider $l_{2}$ norm-ball constrained adversarial attacks as our threat~\cite{goodfellow2014explaining}.\\ 
\heading{Notations.} Noise and noise-perturbed images are represented as $\eta$ and $x^{\ast}$, respectively. We denote our proposed learnable RDUNet as $\mathrm{D}_{\Theta}^{u}$. We denote encoder and decoder as $E_{\Theta_{e}}$ and $D_{\Theta_{d}}$ respectively. We represent our black-box predictor as $f$. We denote RGE  and CGE ZO optimization as R and C, respectively.

\subsection{Proposed Robust Architecture}
We provide a robust black-box defense by pre-pending our proposed robust denoiser RDUNet, followed by a custom-trained AE to the black-box predictor. We show the architecture of our proposed RDUNet in Figure~\ref{fig:fig_arch}. The network has a feedforward and a feedback path. We have a stack of conv layers, and each conv layer contains a convolution layer with kernel size $3 \times 3$ and padding of 1, followed by batch normalization layer~\cite{ioffe2015batch} and rectified linear unit~\cite{krizhevsky2012imagenet}. DConv represents two consecutive conv layers. Our feedforward path consists of five blocks: one is DConv block, and four are Maxpooling + DConv. 

RDUNet has four blocks with a fusion module followed by DConv operation in the feedback path. The last block in the feedback path is a convolution layer with kernel size $1 \times 1$. Fusion module receives two inputs, one feedback input from the feedback path and the lateral input from the feedforward path. We use ConvTranspose2D~\cite{DBLP:journals/corr/DumoulinVL16} to upsample the feedback input to the same size as the lateral input. The feedforward path generates feature maps of increasingly lower resolutions, and along the feedback path, the feature maps have an increasingly higher resolution, as is visible from Figure~\ref{fig:fig_arch}. The denoised output $\hat{x}$ is the sum of noisy image $x^{\ast}$ and output of our proposed arcitecture RDUNet.
\subsection{Proposed Objective Function} Our proposed robustification model is trained by using different losses designed for ensuring three objectives, (i) correct predictions by $f$ on the denoised output $\hat{x}$, (ii) the similarity between the features of clean training samples $x$ and denoised output $\mathrm{D}^{u}_{\Theta}(x^{\ast})$, and (iii) decrease in the domain gap between the probability distributions of clean samples and denoised output.  \\
\heading{Robust Prediction.} We use cross-entropy loss $\mathcal{L}_{CE}$~\cite{zhong2018camera} to make sure that the label predicted by the black-box model of the original input sample is same as that of the denoised output of its Gaussian-perturbed counterpart. 
\begin{equation}
\mathcal{L}_{CE}(\Theta) = \mathbb{E}[-p(f(x)^{l})^{\top}\log(p(f(\mathrm{D}_{\Theta}^{u}(x^{\ast}))^{l}))],
\label{eqn:Equation10}
\end{equation}
where $f(x)$ is a black-box predictor which takes an input $x$ and makes a predictions. $p(f(x)^{l})$ and $p(f(\mathrm{D}_{\Theta}^{u}(x^{\ast}))^{l})$ are the probabilities predicted by $f$ for clean samples and denoised output. 

\heading{Feature Similarity.} We leverage the information that the black-box model trained on the training dataset is highly discriminative. Thus, we propose cosine similarity to learn a mapping between the logit features of the original input images $p(f(x))$ and the logits of the output obtained from the denoiser of the Gaussian perturbed inputs $p(f(\mathrm{D}_{\Theta}^{u}(x^{\ast})))$.
\begin{equation}
\mathcal{L}_{CS}(\Theta) = \mathbb{E}[\frac{p(f(x))^{\top} p(f(\mathrm{D}_{\Theta}^{u}(x^{\ast})))}{\|p(f(x))\|\|p(f(\mathrm{D}_{\Theta}^{u}(x^{\ast})))\|}].   
\label{eqn:Equation11}
\end{equation}
\heading{Domain Similarity.} In addition to maintaining the label and feature consistency at the sample level, we want to bring the domain distribution of synthesized denoised images closer to the original input sample by using maximum mean discrepancy ($MMD(\mu,v)$)~\cite{gretton2012kernel}  on the features of input images $(f(x))$ and features of denoised output $(f(\mathrm{D}_{\Theta}^{u}(x^{\ast})))$. It is given as~\cite{gretton2012kernel},
\begin{equation}
    MMD(\mu,v) = \|\textstyle\sum_{i} \phi(p_{i}) - \textstyle\sum_{i} \phi(q_{i})\|_{\mathcal{H}}^{2},
\end{equation}
where $\| \cdot \|$ is the norm, $\phi$ is a function that maps datapoints to a kernel Hilbert space (RKHS) $\mathcal{H}$, $\{p_{i}\}$ and $\{q_{i}\}$ are samples drawn from distributions $\mu$ and $v$, respectively.

MMD measures the distance between the expected feature map of the samples from the distributions $\mu$ and $v$ in the RKHS ($\mathcal{H}$) induced by $\phi$. Minimization of MMD between between distributions of clean images and denoised output ensures reconstruction of denoised output that are as close as possible to the clean images. Therefore, the utilization of MMD leads to robust training of the denoiser model which when pre-pended to the black-box target model provides certified black-box defense.

This distribution pulling of the original samples and denoised output is inspired by the task of domain adaptation~\cite{mekhazni2020unsupervised,zhang2020discriminative}.
\begin{equation}
\mathcal{L}_{MMD}(\Theta) = MMD(f_{feat}(x), f_{feat}(\mathrm{D}_{\Theta}^{u}(x^{\ast}))).
\label{eqn:Equation12}
\end{equation}

\heading{Our Overall Objective Function.} We optimize our certified defense mechanism with loss functions given in Eq.~\ref{eqn:Equation10},~\ref{eqn:Equation11}~and~\ref{eqn:Equation12} 
\begin{equation}
\mathcal{L}_{Tot}(\Theta) = \mathcal{L}_{CE} + \lambda_{CS}\mathcal{L}_{CS}+\lambda_{MMD}\mathcal{L}_{MMD},
\label{eqn:Equation13}
\end{equation}
where $\lambda_{CS}$ and $\lambda_{MMD}$ are the weights assigned to the loss functions. However, since we cannot access the parameters or weights of the model, we cannot optimize our model using standard optimizers like SGD~\cite{amari1993backpropagation} or ADAM~\cite{zhang2018improved} as that would require back-propagation through the predictor. Thus, we utilize ZO optimization approaches where values of functions are approximated instead of using true gradients.

\subsection{Proposed Black-Box Defense Methods} 
We discuss in this section our two defense methods in detail:\\
\heading{ZO Robust UNet Denoised Smoothing (ZO-RUDS) Defense (RGE Optimization).}
We represent $\mathcal{L}_{Tot}$ as $\mathcal{L}_{Tot}^{R}(\Theta)$ when we optimize our objective function (Eq.~\ref{eqn:Equation13}) with RGE ZO optimization (R)~\cite{liu2020primer}. We calculate gradient estimate of $\mathcal{L}_{Tot}^{R}(\Theta)$ as,
\begin{equation}
     \hat{\nabla}_{\Theta}\mathcal{L}_{Tot}^{R}(\Theta)
     \approx \frac{\delta \mathrm{D}_{\Theta}^{u}(x^{\ast})}{\delta \Theta} \hat{\nabla}_{z}f(z)|_{z=\mathrm{D}_{\Theta}^{u}(x^{\ast})},
\label{eqn:Equation17}
\end{equation}
where $\hat{\nabla}_{z}f(z)$ is the ZO gradient estimate of $f$. We calculate the RGE ZO gradient estimate of $\mathcal{L}_{Tot}^{R}(\Theta)$ by the difference of two function values along a set of random direction vectors. It is represented as:
\begin{equation}
    \begin{split}
        \hat{\nabla}_{\Theta} \mathcal{L}_{Tot}^{R}(\Theta) & = \sum_{k=0}^{q-1}[\frac{d}{\xi\cdot q}(\mathcal{L}_{Tot}(\Theta + \xi u_{k}) - \mathcal{L}_{Tot}(\Theta))u_{k}], \\
    \end{split}
    \label{eqn:Equation14}
\end{equation}
where $d$ represents dimensionality, $q$ is a hyperparameter which represents the querying directions, and $u \in \{1,2,...,q\}$ are $q$ independently and uniformly drawn random vectors from a unit Euclidean sphere. $\xi > 0$ is the smoothing parameter with a small step size of $0.005$. We show the corresponding algorithm of our ZO-RUDS defense in Alg.~\ref{alg:algorithm1}. Zhang~\etal~\cite{zhang2022robustify} directly applied ZO to the previous approach~\cite{salman2020denoised}, with poor performance in the RGE optimization approach. However, we show in Table~\ref{tab:SOTA} in section~\ref{subsec:SOTA}, that after using our proposed RDUNet in ZO-RUDS defense mechanism, we achieve a huge increase of $\textbf{35\%}$ in certified accuracy compared to~\cite{zhang2022robustify}. This proves that our proposed RDUNet type of architecture enables the model to learn fine-scale information while maintaining a low reconstruction error in comparison to previous custom-trained denoisers~\cite{salman2020denoised,zhang2022robustify}.  
\begin{algorithm}
\begin{algorithmic}[1]
\REQUIRE Input $x$, noise $\eta$, smoothing parameter $\xi$, query directions $q$, dimensionality $d$, black-box predictor $f$, initial parameters $\Theta$ of RDUNet.
\ENSURE Trained RDUNet $\mathrm{D}_{\Theta}^{u}$
\STATE $\hat{x} = \mathrm{D}_{\Theta}^{u}(x+\eta) = \mathrm{D}_{\Theta}^{u}(x^{\ast})$,
\STATE Calculate $\mathcal{L}_{Tot}(\Theta)$ (Eq.~\ref{eqn:Equation13}),
\FOR{$k=0$ to $q-1$}
\STATE Obtain a random direction vector $u_{k}$ with Normal distribution $N(\mu,\Sigma)$,
\STATE Calculate $\hat{x}_{q} = \hat{x}+ \xi\cdot u_{k}$,
\STATE Calculate $\mathcal{L}_{Tot}(\Theta)$ (Eq.\ref{eqn:Equation13}),
\STATE Calculate gradient estimation using Eq.\ref{eqn:Equation14},
\ENDFOR
\caption{ZO-RUDS Defense (RGE)}
\label{alg:algorithm1}
\end{algorithmic}
\end{algorithm}

\heading{ZO Autoencoder-based Robust UNet Denoised Smoothing (ZO-AE-RUDS).}
We further pre-pend RDUNet followed by an encoder $E_{\Theta_{e}}$ and a decoder $D_{\Theta_{d}}$ to the black-box predictor for better performance on datasets like Tiny Imagenet and STL-10 with large dimensionality. It ensures that ZO optimization can be carried out in a feature embedding space with low dimensions. However, the autoencoder can also lead to over-reduced features for these datasets leading to poor performance as in~\cite{zhang2022robustify}. Thus, our ZO-AE-RUDS with RDUNet denoiser overcomes this disadvantage, and due to the ability of RDUNet to learn fine-scaled information leads to high performance. The objective function for our ZO-AE-RUDS defense is represented as,
\begin{equation}
\begin{split}
        \mathcal{L}_{Tot}^{C}(\Theta) := \mathcal{L}_{Tot}(f(D_{\Theta_{d}}(z)); z=E_{\Theta_{e}}(\mathrm{D}_{\Theta}^{u}(x^{\ast})),
\end{split}
\end{equation}
where $\mathcal{L}_{Tot}^{C}(\Theta)$ represents that we optimize $\mathcal{L}_{Tot}$ (Eq.~\ref{eqn:Equation13}) with CGE ZO optimization (C)~\cite{liu2020primer}. We calculate gradient estimate of $\mathcal{L}_{Tot}^{C}(\Theta)$ as,
\begin{equation}
     \hat{\nabla}_{\Theta}\mathcal{L}_{Tot}^{C}(\Theta)
     \approx \frac{dE_{\Theta_{e}}(\mathrm{D}_{\Theta}^{u}(x^{\ast}))}{d\Theta} \hat{\nabla}_{z}f(z)|_{z=E_{\Theta_{e}}(\mathrm{D}_{\Theta}^{u}(x^{\ast}))},
\label{eqn:Equation17}
\end{equation}
where $z$ is the latent feature vector with reduced dimension $d_{r} < d$. This reduction in dimension makes the CGE ZO optimization approach feasible. Thus, we utilize CGE~\cite{lian2016comprehensive,liu2018zeroth} and  calculate the gradient estimate of training objective function $\mathcal{L}_{Tot}(\Theta)$ as,
\begin{equation}
    \hat{\nabla}_{\Theta} \mathcal{L}_{Tot}^{C}(\Theta) = \sum_{k=0}^{d-1}[\frac{(\mathcal{L}_{Tot}(\Theta + \xi e_{k}) - \mathcal{L}_{Tot}(\Theta- \xi e_{k}))}{\xi }e_{k}],
    \label{eqn:Equation15}
\end{equation}
where $e^{k} \in \mathbb{R}^{d}$ is the $k$th elementary basis vector, with $1$ at the $k$th coordinate and $0$s elsewhere. We show the corresponding algorithm of our ZO-AE-RUDS defense using CGE optimization in Alg.~\ref{alg:algorithm2}.

\begin{figure}
    \centering
    \includegraphics[width=\linewidth]{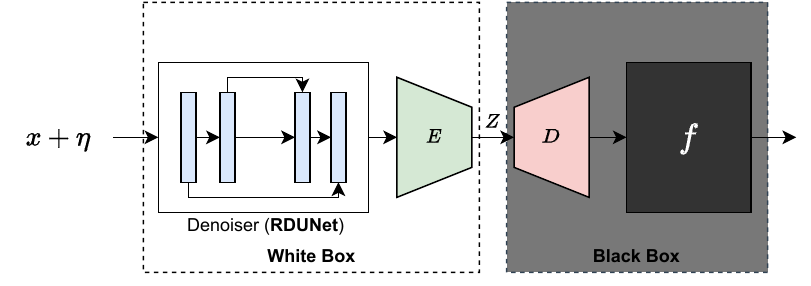}
    \caption{Architecture of our defense technique ZO-AE-RUDS. The decoder `D' and the target model `f'constitute the Black-Box architecture, and therefore, their parameters are not learnt.}
    \label{fig:AE_RUDS}
\end{figure}

In order to tackle with difficulty with ZO optimization in high-dimension datasets like Tiny Imagenet and STL-10, we pre-pend RDUNet followed by autoencoder (AE) to the target predictor. We show our proposed framework for ZO-AE-RUDS in Figure~\ref{fig:AE_RUDS}. We extend our defensive operation where the new black-box is $D+f$, and the new white-box system is RDUNet + $E$ by plugging AE between our RDUNet and black-box predictor $f$.

We observe from Figure~\ref{fig:AE_RUDS} that for low-dimension datasets like CIFAR-10, there is an increase of approximately $2\%$, whereas, for high-dimension dataset STL-10, there is an increase of approximately $10\%$ between performances of ZO-RUDS and ZO-AE-RUDS. Thus, AE enables us to conduct ZO optimization in a feature-embedding space which ensures the feasibility of least-variance CGE. 
\begin{algorithm}
\begin{algorithmic}[1]
\REQUIRE Input $x$, noise $\eta$, smoothing parameter $\xi$, query directions $q$, dimensionality $d$, black-box predictor $f$, $\mathcal{D}_{\Theta_{d}}$ decoder, initial parameters $\Theta$ of RDUNet, and $\Theta_{e}$ of white-box encoder $E_{\Theta_{e}}$
\ENSURE Trained $\mathrm{D}_{\Theta}^{u} +  E_{\Theta_{e}}$  
\STATE  $z = E_{\Theta_{e}}(\mathrm{D}_{\Theta}^{u}(x+\eta))$,
\STATE $\hat{x} = \mathrm{D}_{\Theta}^{u}(z)$,
\STATE Calculate $\mathcal{L}_{Tot}(f(x),f(\hat{x}))$ (Eq.~\ref{eqn:Equation13}),
\FOR{$k=0$ to $d-1$}
\STATE  Obtain a elementary basis direction vector $e_{k}$, 
\STATE Calculate $\hat{x}_{q}^{+} = \hat{x}+ \xi\cdot e_{k}$ and $\hat{x}_{q}^{-} = \hat{x}- \xi\cdot e_{k}$,
\STATE Calculate $\mathcal{L}_{Tot}(f(x),f(E_{\Theta_{e}}(\hat{x}_{q}^{+})))$ (Eq.~\ref{eqn:Equation13}),
\STATE Calculate $\mathcal{L}_{Tot}(f(x),f(E_{\Theta_{e}}(\hat{x}_{q}^{-})))$  (Eq.~\ref{eqn:Equation13}),
\STATE Calculate gradient estimation using Eq.~\ref{eqn:Equation15},
\ENDFOR
\caption{ZO-AE-RUDS Defense (CGE)}
\label{alg:algorithm2}
\end{algorithmic}
\end{algorithm}

\section{Experimental Settings}
\heading{Datasets and Models.} We evaluate the results on CIFAR-10~\cite{krizhevsky2009learning}, CIFAR-100~\cite{krizhevsky2009learn}, Tiny Imagenet~\cite{liu2017tiny} and STL-10~\cite{coates2011analysis} datasets for classification task. For image reconstruction, we focus on MNIST~\cite{lecun1998mnist} dataset. We consider pre-trained models ResNet-110 for CIFAR-10 and ResNet-18 for STL-10. For CIFAR-100, we use Resnet-50 as the target classifier; for Tiny Imagenet dataset, we use ResNet-34. 

\heading{Implementation Details.} We use learning rate $10^{-4}$ and weight decay by 10 at every 100 epochs with total 600 epochs. We set the smoothing parameter $\zeta = 0.005$ for a fair comparison with SOTA. We sample noise $\eta$ with mean $\mu = 0$ and variance $\sigma^{2} = 0.25$ from Normal distribution. We optimize ZO-RUDS with RGE (R) and ZO-AE-RUDS with CGE (C) optimization unless otherwise stated. $\lambda_{MMD}=4$ and $\lambda_{CS}=1$. We use batch-size of 256. We are utilizing a labeled dataset in our work, but our approach is utilizing a black-box model, which means we do not use the labels during processing. Thus, it's important to mention that we do not rely on the original labels of the datasets. This approach enhances the practicality and scalability of our technique, as manually annotating large-scale datasets is not feasible. By bypassing the need for original labels, our method can be efficiently scaled and applied to extensive data without the constraints of traditional labeling requirements. We measure the robustification using standard certified accuracy (SCA) (\%) and robust certified accuracy (RCA) (\%). In each tabular result, we have shown best results for certified black-box defense in bold. We also show the results of certified defense in white-box setting and prove that our results are comparative or better than previous white-box certified defense methods even with no information of model weights or parameters.\\
\textbf{Computational Complexity.} We optimize our training time with the help of parallel processing and matrix operations. The averaged training time on NVIDIA A100-SXM4 for one epoch is approximately $\sim$ 30sec for our first-order (FO) approach FO-AE-RUDS in white-box setting. For our proposed certified black-box defense approach ZO-RUDS (RGE) the averaged training time is $\sim$ 30min and $\sim$ 33min for ZO-AE-RUDS (CGE), on the CIFAR-10 dataset.

\heading{Evaluation Metrics.} We measure the robustification of our model using standard certified accuracy (SCA (\%)) at $l_{2}$-radius $(r)=0$ and robust certified accuracy (RCA (\%)) at $r=\{0.25,0.50,0.75\}$. Higher certified accuracy (CA) ensures that for a given $r$, more percentage of correctly predicted samples have certified radii larger than $r$~\cite{cohen2019certified}.

\subsection{Comparison with SOTA}
\label{subsec:SOTA}
We compare our proposed defense methods ZO-RUDS and ZO-AE-RUDS with previous certified defense approaches in white-box and black-box settings in Table~\ref{tab:SOTA}. In white-box setting we compare our approach with RS~\cite{cohen2019certified}, DS (FO-DS)~\cite{salman2020denoised} and FO-AE-DS~\cite{zhang2022robustify}. FO-AE-DS is the first-order implementation of ZO-AE-DS. We show the results for FO-RUDS and FO-AE-RUDS with our proposed robust denoiser RDUNet in white-box setting. We observe that our proposed FO-RUDS and FO-AE-RUDS shows a performance improvement of approaximately $7\%$ and $3\%$ on FO-DS~\cite{salman2020denoised} and FO-AE-DS~\cite{zhang2022robustify} resepctively, in standard certified accuracy (SCA) and similarly provides robustness for all radii. We show our black-box defense techniques (ZO-RUDS, ZO-AE-RUDS (R), and ZO-AE-RUDS (C)) achieve comparative performance to our proposed defenses in white-box settings (FO-RUDS and FO-AE-RUDS).

Our proposed defense methods ZO-RUDS and ZO-AE-RUDS with RDUNet comfortably outperform SOTA~\cite{zhang2022robustify} by a large margin of $\textbf{35\%}$ and $\textbf{9\%}$ in RGE and CGE optimization approaches respectively. It consistently achieves higher certified robustness across different $r$. The high performance of ZO-RUDS over SOTA signifies that our method leads to an effective DS-oriented robust defense even without additional custom-trained autoencoder. We observe that our method provides better results in the black-box setting and better performance than RS~\cite{cohen2019certified} and DS~\cite{salman2020denoised}, which defend the model in the white-box setting, thus proving the effectiveness of our approach. 

\tabcolsep=3.5pt
\begin{table}[h]
    \centering
    \begin{tabular}{c|c|c|c|c|c}
    \toprule
         \multirow{2}{*}{Method} & \multirow{2}{*}{Type}& \multirow{2}{*}{SCA}&\multicolumn{3}{c}{RCA ($r$)} \\
         \cline{4-6}
          & & & 0.25 & 0.50 & 0.75 \\
         \hline
        RS~\cite{cohen2019certified} & W &76.22 &61.20 &43.23 & 25.67\\
        \hline
        DS (FO-DS)~\cite{salman2020denoised} & W &70.80 &53.31 &40.89 &25.90 \\
        \hline
        \textbf{FO-RUDS} & W& \textbf{77.32} & \textbf{61.78} & \textbf{49.43} & \textbf{34.56}\\
        \hline
        FO-AE-DS~\cite{zhang2022robustify} & W & 75.92 & 60.54 & 46.45 & 32.19 \\
        \hline
         \textbf{FO-AE-RUDS} & W &\textbf{79.95} &\textbf{63.24} &\textbf{49.21} &\textbf{35.82}\\
        \hline
        DS~\cite{salman2020denoised} & B & 74.89&44.56 & 18.20& 14.39\\
        \hline
        ZO-DS (R)~\cite{zhang2022robustify}& B & 42.34&18.12 & 5.01& 0.19\\
        \hline
        \textbf{ZO-RUDS (R)} & B & \textbf{77.89} & \textbf{58.92}& \textbf{38.31} & \textbf{21.93}\\
        \hline
         ZO-AE-DS (R)~\cite{zhang2022robustify} & B& 60.90 & 43.25&26.23 &7.78 \\ 
         \hline
        ZO-AE-DS (C)~\cite{zhang2022robustify} & B &70.90 &53.45 &33.21 & 12.45\\
        \hline
         \textbf{ZO-AE-RUDS (R)} & B &\textbf{76.87} &\textbf{58.23} &\textbf{37.72} &\textbf{20.65}\\
         \hline
        \textbf{ZO-AE-RUDS (C)} & B& \textbf{79.87} & \textbf{61.32}& \textbf{42.90} & \textbf{23.21}\\
        \hline
    \end{tabular}
    \caption{Comparison with SOTA certified defense techniques in white-box (W) and black-box (B) settings on CIFAR-10 dataset. `R' and `C' are RGE and CGE optimization techniques. $`q'=192$.}
    \label{tab:SOTA}
\end{table}
\begin{figure*}[h]
\begin{center}
\includegraphics[width=\linewidth]{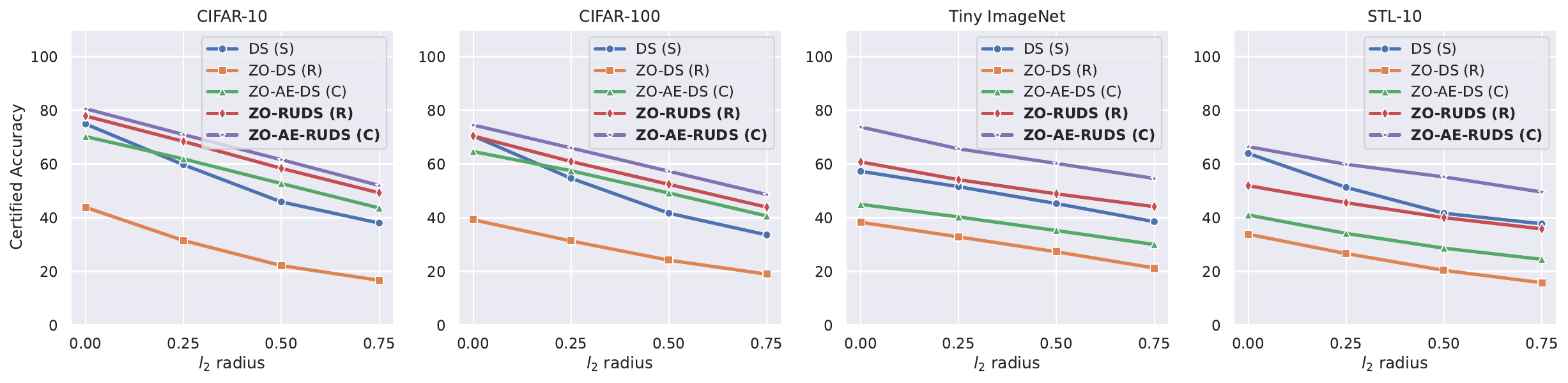}
\caption{Comparison of Certified Accuracy on low-dimension (CIFAR-10, CIFAR-100) and high-dimension (STL-10 and Tiny Imagenet) datasets for different $l_{2}$ radius at query $`q'=192$. `R'- RGE and `C'- CGE ZO techniques.}
\label{Figure_comp}
\end{center}
\end{figure*}
\tabcolsep=3.5pt
\begin{table}[h]
    \centering
    \begin{tabular}{c|c|c|c|c|c}
    \toprule
         \multicolumn{6}{c}{CIFAR-10} \\
         \hline
         q & \multirow{2}{*}{Model}& \multirow{2}{*}{SCA}&\multicolumn{3}{c}{RCA ($r$)} \\
         \cline{4-6}
          & & & 0.25 & 0.50 & 0.75 \\
         \hline
         \multirow{4}*{20} & ZO-DS (R)~\cite{zhang2022robustify} &18.56&3.88 &0.53 &0.26 \\
         & ZO-AE-DS (C)~\cite{zhang2022robustify} &40.67 &27.90 &17.84 &7.10 \\
         & \textbf{ZO-RUDS} (R) &\textbf{62.10} & \textbf{42.43} & \textbf{36.23}&\textbf{24.56} \\
          & \textbf{ZO-AE-RUDS} (C) &\textbf{63.59} & \textbf{51.34} & \textbf{39.01}& \textbf{30.23} \\
         \hline
         \multirow{4}*{100} & ZO-DS (R)~\cite{zhang2022robustify} &39.82 &17.90 &4.71 &0.29 \\
         & ZO-AE-DS (C)~\cite{zhang2022robustify} & 54.32&40.90 &23.98 &9.35\\
         & \textbf{ZO-RUDS} (R) &\textbf{74.60} &\textbf{58.71} & \textbf{39.56}&\textbf{27.82}\\
         & \textbf{ZO-AE-RUDS} (C) &\textbf{76.34} &\textbf{62.46} & \textbf{44.29} &\textbf{30.22} \\
         \hline
         \multicolumn{6}{c}{STL-10} \\
         \hline
         \multirow{4}*{576} & ZO-DS (R)~\cite{zhang2022robustify} &37.59 &21.23 &8.67 &2.56 \\
         & ZO-AE-DS (C)~\cite{zhang2022robustify} &44.78 &33.41 &26.10 &16.43 \\
         & \textbf{ZO-RUDS} (R) & \textbf{58.20} &\textbf{47.83} &\textbf{40.32} &\textbf{29.89} \\
          & \textbf{ZO-AE-RUDS} (C) &\textbf{68.29} &\textbf{57.93} & \textbf{47.31}&\textbf{33.22} \\
         \bottomrule
    \end{tabular}
    \caption{Comparison of our defense with previous ZO defense approaches.  }
    \label{tab:CIFAR10_stl10}
\end{table}
\begin{table*}[bp]
    \centering
    \begin{tabular}{c|c c|c c|c c|c c|c c}
    \toprule
         \multirow{2}*{Method} & \multicolumn{2}{c|}{$\|\delta||_{2}=0$} & \multicolumn{2}{c|}{$\|\delta||_{2}=1$} & \multicolumn{2}{c|}{$\|\delta||_{2}=2$} & \multicolumn{2}{c|}{$\|\delta||_{2}=3$} & \multicolumn{2}{c}{$\|\delta||_{2}=4$} \\
         \cline{2-11}
         & RMSE & SSIM & RMSE & SSIM & RMSE & SSIM & RMSE & SSIM & RMSE & SSIM\\
         \hline
         Vanilla~\cite{raj2020improving} & 0.1213 & 0.7934 & 0.3251 & 0.4367 & 0.4629 & 0.1468 & 0.6129 & 0.04945 & 0.5976 & 0.0168 \\
         FO-DS~\cite{salman2020denoised} & 0.1596 & 0.7415 & 0.1692 & 0.6934 & 0.2182 & 0.5421 & 0.2698 & 0.3956 & 0.3245 & 0.3178 \\
         FO-AE-DS~\cite{zhang2022robustify} & 0.1475 & 0.7594 & 0.1782 & 0.7025 & 0.2182 & 0.5421 & 0.2693 & 0.4163 & 0.3176 & 0.3293 \\
         ZO-DS (R)~\cite{zhang2022robustify} & 0.1892 & 0.5345 & 0.2267 & 0.4634 & 0.2634 & 0.3689 & 0.3092 & 0.2792 & 0.3482 & 0.2177 \\
         ZO-AE-DS (C)~\cite{zhang2022robustify} & 0.1398 & 0.6894 & 0.1634 & 0.7099 & 0.2126 & 0.5472 & 0.2689 & 0.4188 & 0.3367 & 0.3294 \\
         \textbf{ZO-RUDS} (R) & \textbf{0.1232} & \textbf{0.7924} & \textbf{0.1465} & \textbf{0.7991} & \textbf{0.2053} & \textbf{0.5966} & \textbf{0.2380} & \textbf{0.4591} & \textbf{0.3082} & \textbf{0.3811} \\
         \textbf{ZO-AE-RUDS} (C) & \textbf{0.1219} & \textbf{0.7926} & \textbf{0.1392} & \textbf{0.8346} & \textbf{0.1872} & \textbf{0.6648} & \textbf{0.2174} & \textbf{0.6102} & \textbf{0.2679}& \textbf{0.5236} \\
         \bottomrule
    \end{tabular}
    \caption{Performance comparison of image reconstruction task on MNIST dataset. ($`q'=192$) }
    \label{tab:recon}
\end{table*}

\subsection{ Performance on Image Classification}
\label{subsec:Image}
\heading{Performance on different number of queries.} We show the performance for other queries $q={20,100}$ for CIFAR-10 and $q={576}$ for STL-10 dataset in Table~\ref{tab:CIFAR10_stl10}. Our proposed robustification outperforms SOTA~\cite{zhang2022robustify} on all these queries by a huge margin for SCA and RCA evaluation metrics at different $l_{2}$ radii. We observe that~\cite{zhang2022robustify} has poor performance for high-dimension images even after increasing the number of queries or using CGE optimization with auto-encoder to decrease the variance caused by RGE optimization. This may happen due to two reasons. First, the bottleneck in the autoencoder architecture constrains the fine-scaled information necessary for reconstructing denoised images. Second, the over-reduced feature dimension in high-dimension images could hamper the performance. Our proposed denoiser RDUNet with lateral connections between the encoder and decoder ensures better reconstruction of denoised output, which ensures prediction with high certified accuracy. RDUNet decreases model variance as it consists of downsampling and upsampling layers in the encoding and decoding path, which makes the model invariant to changes in image dimensions and thus performs better for high dimensions as well~\cite{peng2019densely,badrinarayanan2015segnet}. The lateral skip connections help the model learn fine-scale information, thus overcoming the disadvantage of auto-encoder constraining fine-scale information.

\heading{Performance on low-dimension (CIFAR-10, CIFAR-100) and high-dimension (STL-10, Tiny Imagenet) classification datasets.} We compare our proposed defense methods ZO-RUDS (R) and ZO-AE-RUDS (C) with previous black-box defense methods DS (S)~\cite{salman2020denoised}, ZO-DS (R)~\cite{zhang2022robustify} and ZO-AE-DS (C)~\cite{zhang2022robustify}  at various $r$ for low and high-dimension classification datasets in Figure~\ref{Figure_comp}. Our proposed approaches ZO-RUDS and ZO-AE-RUDS beat the SOTA~\cite{zhang2022robustify} by a large margin of $\textbf{30.21\%}$ and $\textbf{8.87\%}$ in RGE, and CGE optimization approaches respectively for CIFAR-100 dataset. It outperforms SOTA by a huge margin for all other radii as well. We show that unlike~\cite{zhang2022robustify}, which gives better performance only for the CGE optimization approach after appending autoencoder in ZO-AE-RUDS, our proposed RDUNet denoiser when appended to predictor performs better for both RGE and CGE optimization approaches. We observe that SOTA~\cite{zhang2022robustify} fails to perform even after the addition of an autoencoder to the network for high-dimension Tiny Imagenet and STL-10 datasets. Our defense methods ZO-RUDS and ZO-AE-RUDS beats SOTA~\cite{zhang2022robustify} by a huge margin of $\textbf{24.81\%}$ and $\textbf{25.84\%}$ respectively for Tiny Imagenet dataset. Similar observations can be made for CIFAR-10 and STL-10 datasets at different certified radii, as shown in Figure~\ref{Figure_comp}.

\subsection{Performance on Image Reconstruction} 
Previous works~\cite{antun2020instabilities,raj2020improving,wolfmaking} show that image reconstruction networks are vulnerable to adversarial attacks like PGD attacks~\cite{madry2017towards}. We compare our proposed defense methods ZO-RUDS and ZO-AE-RUDS with previous white-box and black-box defense methods in Table~\ref{tab:recon}. We follow the settings of~\cite{zhang2022robustify} and aim to recover the original sample using a pre-trained reconstruction network~\cite{raj2020improving} under adversarial perturbations generated by a 40-step $l_{2}$ PGD attack under $||\delta||_{2}=\{0,1,2,3,4\}$. We use root mean square error (RMSE) and structural similarity (SSIM)~\cite{hore2010image} to find the similarity between the original and reconstructed image. We observe that our method beats SOTA~\cite{zhang2022robustify} with low RMSE and high SSIM scores for all the values of $||\delta||_{2}$. Our defense beats FO defense methods as well as the vanilla model, proving the robustness provided by our method for the reconstruction task. We observe that at high values of $||\delta||_{2}$ perturbation the performance of vanilla model~\cite{raj2020improving} decreases drastically.
\tabcolsep=3pt
\begin{table}[h]
    \centering
    \begin{tabular}{c|c|c|c|c}
    \toprule
         \multirow{2}{*}{Loss} & \multirow{2}{*}{SCA}&\multicolumn{3}{c}{RCA ($r$)} \\
         \cline{3-5}
          &   & 0.25 & 0.50 & 0.75 \\
         \hline
         $\mathcal{L}_{CE}$ &72.67 &54.90 & 36.01&17.03 \\
         \hline
         $\mathcal{L}_{CE}+\mathcal{L}_{CS}$ &73.21 &56.04 &36.45 &18.79 \\
         \hline
         $\mathcal{L}_{CE}+\mathcal{L}_{CS}+\mathcal{L}_{MMD}$ & \textbf{79.87}& \textbf{61.32}& \textbf{42.90}& \textbf{23.21} \\
         \bottomrule
    \end{tabular}
    \caption{Effect of loss functions on our proposed (ZO-AE-RUDS) for $`q'=192$. Dataset is CIFAR-10.}
    \label{tab:loss}
\end{table}

\tabcolsep=4pt
\begin{table}[h]
    \centering
    \begin{tabular}{c|c|c|c|c}
    \toprule
         \multirow{2}{*}{Training Strategy} & \multirow{2}{*}{SCA}&\multicolumn{3}{c}{RCA ($l_{2}$-radius)} \\
         \cline{3-5}
          &   & 0.25 & 0.50 & 0.75 \\
         \hline
         $(\mathrm{D}_{\Theta}^{u})_{finetune}$ &67.52 & 53.44&29.56 & 12.87\\
         \hline
         $(\mathrm{D}_{\Theta}^{u})_{scratch}$ &\textbf{79.87} &\textbf{61.32} &\textbf{42.90} &\textbf{23.21} \\
         \bottomrule
    \end{tabular}
    \caption{Effect of RDUNet training strategies on ZO-AE-RUDS for $`q'=192$. Dataset is CIFAR-10.}
    \label{tab:train_strat}
\end{table}
\begin{figure}[t]
\centering
\includegraphics[width=\linewidth]{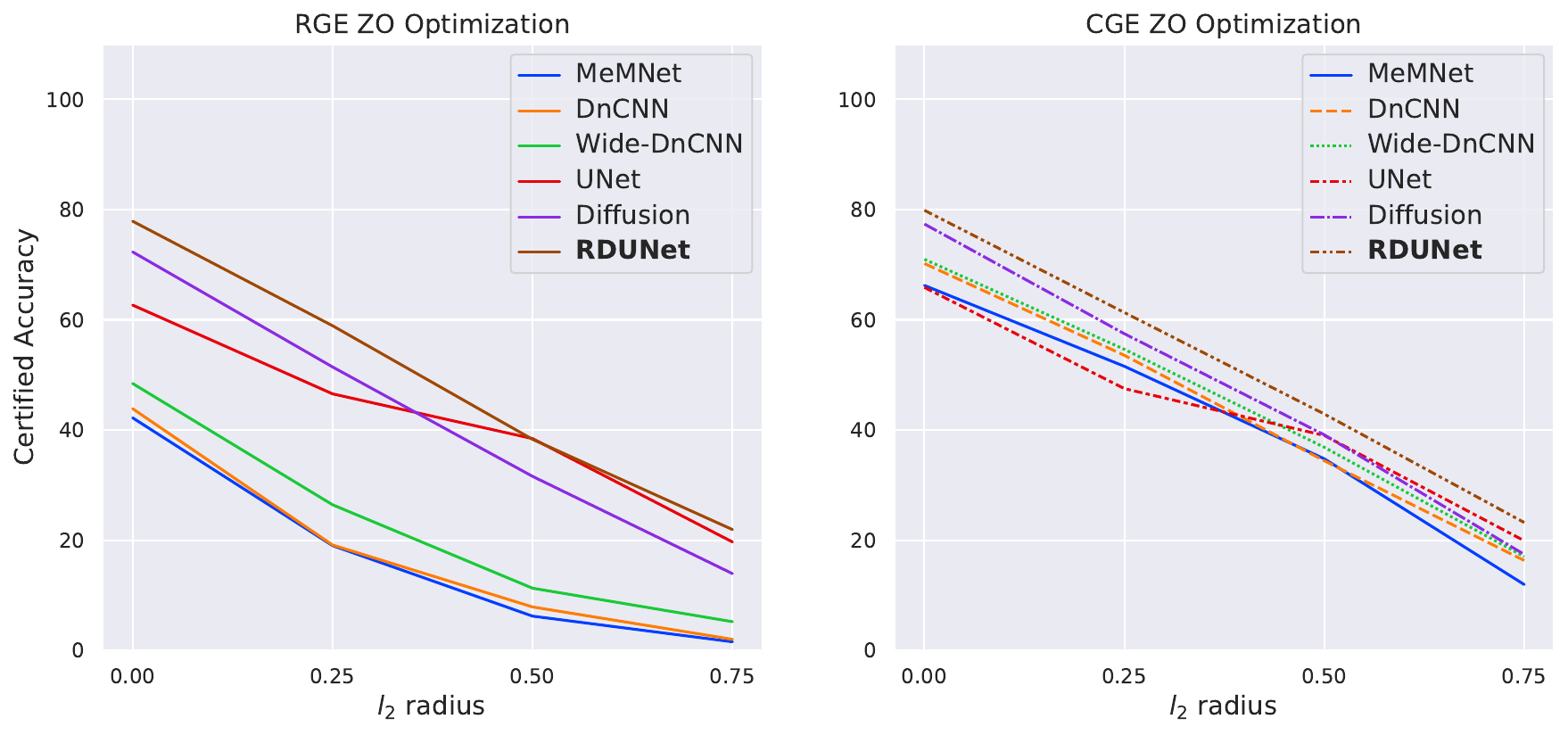}
\caption{Effect of different denoisers on our RGE and CGE ZO optimization-based defense approaches for different $l_{2}$-radii at $`q'=192$. Dataset is CIFAR-10.}
\label{fig:Comp_low_high}
\end{figure}
\subsection{Ablation Study}
We show the effect of loss functions and different denoiser architectures in our proposed defense mechanism.

\heading{Effect of various denoisers.}
We compare our proposed robust denoiser RDUNet with previous denoisers MemNet~\cite{tai2017memnet}, DnCNN~\cite{zhang2017beyond}, wide-DnCNN~\cite{zhang2017beyond}, UNet~\cite{ronneberger2015u} and Diffusion~\cite{ho2020denoising} in Figure~\ref{fig:Comp_low_high}. We show that after appending denoiser DnCNN as proposed in~\cite{salman2020denoised,zhang2022robustify} with 17 layers primarily including Conv+BN+ReLU layers and wide-DnCNN with 128 deep layers to the black-box model in the RGE optimization leads to poor performance. We observe that appending these denoisers to the autoencoder architecture in the CGE optimization improves the robustfication of the model to some extent. We show that the diffusion model gives comparatively better performance than other denoisers. However, RDUNet gives better performance than diffusion models. It may happen due to multiple noise addition, which makes ZO optimization of the diffusion model unstable. These results conclude that our proposed RDUNet provides robustness to a black-box model against adversarial perturbations.

\heading{Effect of Loss Functions.} We show in Table~\ref{tab:loss}, the effect of cosine similarity loss $\mathcal{L}_{CS}$ at the sample level and MMD loss $\mathcal{L}_{MMD}$ between the probability distributions of the original samples and denoised output. We observe that after applying $\mathcal{L}_{CS}$ in addition to cross-entropy loss, our model's performance increases by $1\%$ and using $\mathcal{L}_{MMD}$ the performance increases by approximately $6\%$. This shows that bringing closer the features of the original sample and denoised output at the instance and domain level increase the robustness of the black-box model.\\
\heading{Effect of training strategies.}
We explore how different training strategies affect the performance of RDUNet in the ZO-AE-RUDS model, as shown in Table~\ref{tab:train_strat}. Here, $(\mathrm{D}_{\Theta}^{u})_{finetune}$ means that we first pre-train RDUNet before specifically adapting it to our defense method through further training. On the other hand, $(\mathrm{D}_{\Theta}^{u})_{scratch}$ refers to training RDUNet from the beginning without any pre-training. We show that training from scratch gives better performance than pre-training and fine-tuning RDUNet. We observe during training that $(\mathrm{D}_{\Theta}^{u})_{finetune}$ achieves very high performance at the initial training stage; however, it does not improve as training progresses. Pre-training the denoiser causes the optimization to get stuck at a local optima leading to decreased performance.\\
\heading{Effect of different denoisers.} We calculate model variance for our task of black-box defense with MemNet~\cite{tai2017memnet}, DnCNN~\cite{zhang2017beyond}, Wide-DnCNN~\cite{zhang2017beyond}, UNet~\cite{ronneberger2015u} , Diffusion model~\cite{ho2020denoising} and RDUNet denoiser in Table~\ref{tab:my_label_variance}. We can refer to Eq.7 to calculate model variance. Expanding the loss term $\mathcal{L}_{Tot}(\Theta + \xi u_k)$ using Taylor series and considering only the first order term, we can see that $\hat{\nabla}_{\Theta}\mathcal{L}_{Tot}^{R}(\Theta)$ is equal to $\sum^{(q-1)}_{k=0}[du_k^{\top}\nabla \mathcal{L}_{Tot}(\Theta)u_k/q]$. We compute the model variance using these gradients and observe that RDUNet attains the least variance.
\begin{table}[h]
    \centering
    \small
    \begin{tabular}{p{1.1cm}|p{1cm}|p{1cm}|p{0.7cm}|p{1.1cm}|p{0.9cm}}
    \toprule
         MeMNet  & DnCNN  & Wide-DnCNN  & UNet  & Diffusion  & RDUNet\\
         \hline
         3.4 &3.2 & 3 & 2 & 1.4 & 1 \\
         \bottomrule
    \end{tabular}
    \caption{Model Variance with different denoisers ($\times e-4$).}
    \label{tab:my_label_variance}
\end{table}\\

\begin{figure}[h]
    \centering
    \includegraphics[width=\linewidth]{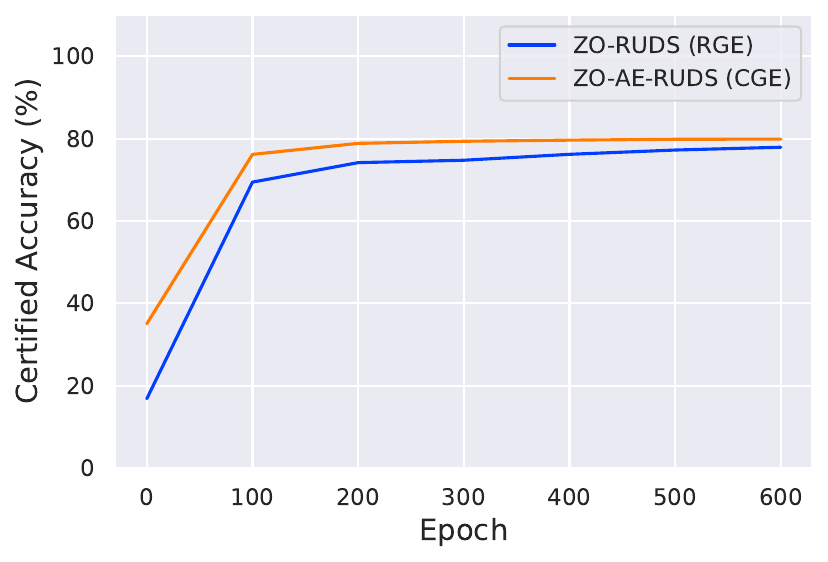}
    \caption{Training progress of our certified black-box defense ZO-RUDS (RGE) and ZO-AE-RUDS (CGE) over number of epochs. Dataset used is CIFAR-10. Query $`q' = 192$, $\sigma = 0.25$.}
    \label{fig:rgecge}
\end{figure}

\heading{Effect of noise level $(\sigma)$.}
In Tables~\ref{tab:CIFAR10_sigma}~and~\ref{tab:stl10_sigma}, we present a comprehensive analysis of the performance of our proposed defense techniques across varying values of $\sigma = {0.50,1.0}$. Our evaluation reveals a consistent and noteworthy superiority of our defense methods over state-of-the-art black-box defense strategies when applied to both the low-dimensional CIFAR-10 and high-dimensional STL-10 datasets across the entire range of $\sigma$ values. Notably, our techniques demonstrate a remarkable ability to maintain resilience against adversarial attacks, showcasing their efficacy even in the presence of heightened noise levels. However, it's worth noting that as the standard deviation of the Gaussian noise added to the input images increases, we do observe a gradual reduction in overall robustness.
\tabcolsep=4pt
\begin{table}[h]
    \centering
    \begin{tabular}{c|c|c|c|c|c}
    \toprule
         \multicolumn{6}{c}{CIFAR-10} \\
         \hline
          & \multirow{2}{*}{Model}& \multirow{2}{*}{SCA}&\multicolumn{3}{c}{RCA ($r$)} \\
         \cline{4-6}
          ($\sigma$)& & & 0.25 & 0.50 & 0.75 \\
         \hline
         \multirow{6}*{0.5} & FO-DS~\cite{salman2020denoised} &66.93 & 45.78&23.34 &6.20 \\
          & ZO-DS (R)~\cite{zhang2022robustify} &35.67 &12.45 &2.58 &0.01 \\
        & FO-AE-DS~\cite{zhang2022robustify} & 67.78&48.90 & 26.76& 9.21\\
         & ZO-AE-DS (C)~\cite{zhang2022robustify} & 63.24&42.98 & 25.67&4.56 \\
         & \textbf{ZO-RUDS (R)} & \textbf{70.44}& \textbf{49.63}& \textbf{31.21}&\textbf{12.48} \\
          & \textbf{ZO-AE-RUDS (C)} &\textbf{71.52} &\textbf{49.68} &\textbf{32.37} &\textbf{13.56} \\
         \hline
         \multirow{6}*{1.0} & FO-DS~\cite{salman2020denoised} &51.29 & 30.47&8.36 &1.67 \\
         & ZO-DS (R)~\cite{zhang2022robustify} & 21.89& 6.34& 0.12& 0.00\\
         & FO-AE-DS~\cite{zhang2022robustify} &54.97 & 37.83& 20.63& 5.64 \\
         & ZO-AE-DS (C)~\cite{zhang2022robustify} &48.90 &37.21 &18.39 &0.79 \\
         & \textbf{ZO-RUDS (R)} & \textbf{55.25} & \textbf{38.98}& \textbf{23.54}&\textbf{3.58}\\
         & \textbf{ZO-AE-RUDS (C)} &\textbf{55.26} &\textbf{38.84} &\textbf{22.49} & \textbf{4.29}\\
         \bottomrule
    \end{tabular}
    \caption{Comparison of our proposed approach with previous ZO optimization approaches on the CIFAR-10 dataset for different noise levels. Query $`q'=192$.}
    \label{tab:CIFAR10_sigma}
\end{table}
\tabcolsep=4pt
\begin{table}[h]
    \centering
    \begin{tabular}{c|c|c|c|c|c}
    \toprule
         \multicolumn{6}{c}{STL-10} \\
         \hline
          & \multirow{2}{*}{Model}& \multirow{2}{*}{SCA}&\multicolumn{3}{c}{RCA ($r$)} \\
         \cline{4-6}
          ($\sigma$)& & & 0.25 & 0.50 & 0.75 \\
         \hline
         \multirow{4}*{0.5} & ZO-DS (R)~\cite{zhang2022robustify} &31.56 & 16.34& 1.57& 0.00\\
         & ZO-AE-DS (C)~\cite{zhang2022robustify} &34.67 & 22.65&18.42 & 9.81\\
                     & \textbf{ZO-RUDS (R)} &\textbf{49.97} &\textbf{39.81} &\textbf{33.54} &\textbf{20.10} \\
                      & \textbf{ZO-AE-RUDS (C)} & \textbf{61.20}& \textbf{48.71}& \textbf{31.23}&\textbf{20.51} \\
         \hline
         \multirow{4}*{1.0} & ZO-DS (R)~\cite{zhang2022robustify} & 12.56& 1.24& 0.34& 0.09\\
         & ZO-AE-DS (C)~\cite{zhang2022robustify} &22.65 &13.42 & 6.32&0.05 \\
         & \textbf{ZO-RUDS (R)} &\textbf{33.28} & \textbf{24.51}& \textbf{16.53}&\textbf{7.62}\\
         & \textbf{ZO-AE-RUDS (C)} & \textbf{43.27}& \textbf{31.27}&\textbf{23.33} &\textbf{9.80} \\
         \bottomrule
    \end{tabular}
    \caption{Comparison of our proposed approach with previous ZO optimization approaches on STL-10 dataset for different noise levels. Query $`q'=192$.}
    \label{tab:stl10_sigma}
\end{table}
\tabcolsep=4pt
\begin{table}[h]
    \centering
    \begin{tabular}{c|c|c|c|c}
    \toprule
         \multicolumn{5}{c}{VGG-16} \\
         \hline
          \multirow{2}{*}{Model}& \multirow{2}{*}{SCA}&\multicolumn{3}{c}{RCA ($r$)} \\
         \cline{3-5}
           & & 0.25 & 0.50 & 0.75 \\
         \hline
         \textbf{ZO-RUDS (R)} &77.24 &59.93 & 38.01& 21.35\\
           \textbf{ZO-AE-RUDS (C)} &79.83 & 61.34& 43.66&23.90 \\
         \hline
         \multicolumn{5}{c}{ViT-16-L(224)~\cite{tseng2022perturbed}} \\
         \hline
          \textbf{ZO-RUDS (R)} &77.91 &60.24 &38.75 &21.26 \\
           \textbf{ZO-AE-RUDS (C)} & 79.89& 62.36&43.87 & 24.01\\
         \bottomrule
    \end{tabular}
    \caption{Certified Accuracy for different classifiers on CIFAR-10 dataset for noise level $\sigma =\{ 0.25\}$. Query $`q'=192$.}
    \label{tab:diff_classifier}
\end{table}\\
\heading{Effect of different classifiers.}
In Table~\ref{tab:diff_classifier}, we've displayed how well our defense techniques ZO-RUDS (R) and ZO-AE-RUDS (C) perform when tested on different classifiers, specifically VGG-16 and Vision Transformer (ViT-16-L(224)). These results clearly demonstrate that our defense method is effective in making a wide range of target models more resilient, regardless of their architecture. So, whether it's VGG-16 or Vision Transformer, our defense techniques consistently bolster their robustness against various types of attacks.\\
\heading{Performance variation over number of epochs.}
We show the RGE and CGE optimization of our black-box defense ZO-RUDS and ZO-AE-RUDS, as training progresses in Figure~\ref{fig:rgecge}. We observe that ZO-AE-RUDS (CGE) gives high performance at initial epochs, however as training progresses both defense approaches give comparative performance. We observe high increase in performance at initial epochs and slight increase in accuracy at higher epochs.
\section{Discussion}
\heading{Training Comparison from previous works.}
We want to clarify that our training method is different from~\cite{zhang2022robustify,chen2017zoo}.~\cite{zhang2022robustify} follows ZOO from~\cite{chen2017zoo} and uses only CE loss $\mathcal{L}_{CE}$ for training. While we also follow ZOO, we propose cosine similarity loss $\mathcal{L}_{CS}$ between features of original input and the denoised output (Eq. 2 in main paper). In addition to feature similarity, we reduce the disparity between the domain distributions of synthesized images and the original input samples by leveraging the MMD $\mathcal{L}_{MMD}$ (Eq. 3 in main paper). We further use $\mathcal{L}_{CE}$ as given in~\cite{zhang2022robustify}.
\tabcolsep=3.5pt
\begin{table}[h]
    \centering
    \begin{tabular}{c|c|c|c|c|c}
    \hline
        \multicolumn{6}{c}{STL-10} \\
        \hline
         \multirow{2}{*}{$q$} & \multirow{2}{*}{Model}& \multirow{2}{*}{SCA}&\multicolumn{3}{c}{RCA ($r$)} \\
         \cline{4-6}
         & & & 0.25 & 0.50 & 0.75 \\
         \hline
        \multirow{2}{*}{576} &ZO-RUDS (R) & 58.20 & 47.83 & 40.32 & 29.89 \\
        &ZO-AE-RUDS (R) & 52.47 & 42.14 &  35.81 & 24.17 \\
         & \textbf{ZO-AE-RUDS} (C) &\textbf{68.29} &\textbf{57.93} & \textbf{47.31}&\textbf{33.22} \\
        \hline
         \multicolumn{6}{c}{CIFAR-10}\\
        \hline
        \multirow{3}{*}{192} & ZO-RUDS (R) & 77.89 & 58.92 & 38.31 & 21.93 \\
        &ZO-AE-RUDS (R) & 76.87 & 58.93 & 37.72 & 20.65 \\
        &\textbf{ZO-AE-RUDS (C)} & \textbf{79.87} & \textbf{61.32}& \textbf{42.90} & \textbf{23.21}\\
        \bottomrule
         \end{tabular}
    \caption{Performance using RGE (R) and CGE (C).}
    \label{tab:my_label}
\end{table}\\
\heading{Importance of RDUNet denoiser in AE+RDUNET architecture.}
In Table 2, we show that when RGE is used, ZO-AE-RUDS (R) gives poor performance compared to ZO-RUDS (R) for both STL-10 and CIFAR-10. The results for STL-10 are worse. This is because STL-10 is a high dimensional dataset compared to CIFAR-10 and applying AE leads to over reduction of features. In contrast, when we use CGE, ZO-AE-RUDS (C), we see that performance is much better. This is because CGE gives lesser model variance [79] and better features. Further, our RDUNet also leads to model variance reduction as shown in Table \ref{tab:my_label_variance}. As CGE acts co-ordinate wise, it is extremely expensive to perform CGE without reducing dimensions with AE and it is necessary to use AE. However, our overall construction leads to better performance as indicated in experiments. \\

\section{Conclusion}
In this work, we study the problem of certified black-box defense, aiming to robustify the black-box model with access only to input queries and output feedback. First, we proposed two novel defense mechanisms, ZO-RUDS and ZO-AE-RUDS, which substantially enhance the defense and optimization performance by reducing the variance of ZO gradient estimates. Second, we proposed a novel robust denoiser RDUNet that provides a scalable defense by directly integrating denoised smoothing with RGE ZO optimization, which was not feasible in previous works. We show that RDUNet gives high performance by further appending autoencoder (AE) as in ZO-AE-RUDS defense. Lastly, we proposed an objective function with MMD loss, bringing the distribution of denoised output closer to clean data. Our elaborate experiments demonstrate that ZO-RUDS and ZO-AE-RUDS achieve SOTA-certified defense performance on classification and reconstruction tasks.  

\bibliographystyle{IEEEtran}
\bibliography{egbib}

\end{document}